\documentclass{article}

\usepackage[preprint,nonatbib]{neurips_2024}

\usepackage[utf8]{inputenc} 
\usepackage[T1]{fontenc}    
\usepackage{hyperref}       
\usepackage{url}            
\usepackage{makecell}
\usepackage{booktabs}       
\usepackage{amsfonts}       
\usepackage{nicefrac}       
\usepackage{microtype}      
\usepackage{xcolor}         
\usepackage{url}
\usepackage{multirow}
\usepackage{colortbl}
\usepackage{tabularx}
\usepackage{CJK}
 \usepackage{graphicx}
 \usepackage{amsmath} 
 
\title{FLAME: Financial Large-Language Model Assessment and Metrics Evaluation}

\author{%
  Jiayu Guo, Yu Guo, Martha Li, Songtao Tan~\thanks{Corresponding Author.}\\
  The School of Finance, Renmin University of China\\
}

\begin{document}
\begin{CJK}{UTF8}{gbsn}

\maketitle

\begin{abstract}
LLMs have revolutionized NLP and demonstrated potential across diverse domains. More and more financial LLMs have been introduced for finance-specific tasks, yet comprehensively assessing their value is still challenging. In this paper, we introduce FLAME, a comprehensive financial LLMs evaluation system in Chinese, which includes two core evaluation benchmarks: \textbf{FLAME-Cer} and \textbf{FLAME-Sce}. FLAME-Cer covers 14 types of authoritative financial certifications, including CPA, CFA, and FRM, with a total of approximately 16,000 carefully selected questions. All questions have been manually reviewed to ensure accuracy and representativeness. FLAME-Sce consists of 10 primary core financial business scenarios, 21 secondary financial business scenarios, and a comprehensive evaluation set of nearly 100 tertiary financial application tasks. We evaluate 6 representative LLMs, including GPT-4o, GLM-4, ERNIE-4.0,  Qwen2.5, XuanYuan3, and the latest Baichuan4-Finance, revealing Baichuan4-Finance excels other LLMs in most tasks. By establishing a comprehensive and professional evaluation system, FLAME facilitates the advancement of financial LLMs in Chinese contexts. Instructions for participating in the evaluation are present on GitHub~\url{ https://github.com/FLAME-ruc/FLAME/}.

\end{abstract}

\section{Introduction}

Large Language Models (LLMs) have revolutionized natural language processing (NLP) and showcased exceptional abilities across specialized domains such as law, finance, and biology~\cite{chen2023chatgpt,dubey2024llama,yang2023baichuan,yang2024qwen2}. In the financial domain, recent studies have demonstrated the considerable potential of advanced LLMs in performing financial tasks. Additionally, many finance-specific LLMs have also been developed.
However, benchmarks for evaluating these LLMs from a professional financial perspective remain to be explored.

Current evaluation benchmarks in the financial domain, such as PIXIU~\cite{xie2023pixiu}, FinBen~\cite{xie2024finben}, FinanceBench~\cite{islam2023financebench} and BizBench~\cite{koncel2023bizbench}, mainly evaluates LLMs from an NLP perspective, such as text similarity, text n-gram overlap and so on. Practical financial applications are seldom considered in these benchmarks. With this motivation in mind, to comprehensively evaluate the capabilities of large language models in the financial domain, we have simultaneously launched the Financial Large-Language Model Assessment and Metrics Evaluation benchmark, called FLAME. FLAME systematically constructs holistic evaluation datasets ranging from financial qualification certifications to real-world application scenarios, providing a reliable assessment standard for developing financial LLMs.

 The FLAME evaluation system comprises two primary assessment benchmarks: FLAME-Cer (Financial Certification) and FLAME-Sce (Financial Scenario). The evaluation system is distinguished by several essential features. 

\begin{itemize}
    \item Developed by the School of Finance at Renmin University of China, it ensures professionalism by merging academic theory with practical experience, thereby establishing the content’s authority and expertise. 
    \item Comprehensiveness is attained by encompassing both knowledge aspects through FLAME-Cer and application aspects through FLAME-Sce, providing a thorough evaluation of models from specialized knowledge to practical implementation. 
    \item The system prioritizes applicability by meticulously aligning evaluation tasks with the genuine needs of the financial industry, concentrating on key application scenarios in banking and insurance, such as product sales, risk management, and customer service. 
    \item Additionally, FLAME rigorously adheres to regulatory compliance standards and assesses models from the perspectives of data security, business compliance, and risk control to ensure their applications conform to financial industry regulations.
\end{itemize}

The FLAME evaluation content is continuously updated to reflect developments in financial markets and regulatory requirements, ensuring the system remains timely and practical. We aim to provide professional, comprehensive, and practical evaluation standards with the FLAME evaluation system.

\section{FLAME}

FLAME comprises two core evaluation benchmarks: the Financial Qualification Certification benchmark (FLAME-Cer) and the Financial Scenario Application benchmark (FLAME-Sce).

\subsection{FLAME-Cer: Financial Qualification Certification Benchmark}

FLAME-Cer is designed to provide a robust evaluation framework by encompassing 14 of the most representative international and domestic financial certifications, consisting of AFP (Associate Financial Planner), CAA (China Actuarial Association), CFA (Chartered Financial Analyst), CIA (Certified Internal Auditor), CISA (Certified Information Systems Auditor), CMA (Certified Management Accountant), CPA (Certified Public Accountant), FRM (Financial Risk Manager), CLIQ (Chinese Life Insurance Qualification), FundPQ (Fund Practitioner Qualification), FuturesPQ (Futures Practitioner Qualification), Economist, SPQ (Securities Practitioner Qualification), and CCBP (Certification of China Banking Professional). 

FLAME-Cer has approximately 16,000 expertly curated questions, ensuring their accuracy and representativeness. Besides, the tasks in benchmark FLAME-Cer are scientifically graded into three significance levels (*, **, ***) from aspects of market recognition, industry influence, and professional depth.
\begin{itemize}
    \item *** represents core industry certification with high market recognition and stringent professional requirements.
    \item ** represents significant professional certification that offers strong practical guidance in the specific field.
    \item * represents auxiliary certification that serves as supplementary professional skills for financial practitioners.
\end{itemize}

The detailed tasks' introductions are as follows:

\begin{itemize}

    \item ** \textbf{AFP (Associate Financial Planner)}: a professional certification in Chinese financial planning, covering basics of financial planning, comprehensive household planning, investment planning, risk management and insurance planning, employee benefits and retirement planning, personal income tax optimization, and more.
    
    \item *** \textbf{CAA (China Actuarial Association)}: a professional qualification exam for actuaries in China, divided into associate actuary and actuary levels, covering actuarial theory, practice, and related laws and regulations.
    
    \item *** \textbf{CFA (Chartered Financial Analyst)}: a globally recognized certification, covering a wide range of financial investment knowledge, including ethics and professional standards, quantitative methods, economics, financial statement analysis, corporate finance, portfolio management, and more.
    
    \item ** \textbf{CIA (Certified Internal Auditor)}: the only globally recognized certification for internal auditing, covering fundamentals, practices, and knowledge elements of internal auditing.
    
    \item * \textbf{CISA (Certified Information Systems Auditor)}: a globally recognized certification for information systems auditors, covering auditing processes, IT governance and management, system acquisition, development and implementation, operations and business continuity, and information asset protection.
    
    \item ** \textbf{CMA (Certified Management Accountant)}: a globally recognized certification in management accounting, covering financial planning, performance analysis, strategic financial management, and more.
    
    \item *** \textbf{CPA (Certified Public Accountant)}: the professional certification exam for Chinese certified public accountants, including six subjects: accounting, auditing, financial cost management, economic law, tax law, and company strategy and risk management.
    
    \item ** \textbf{FRM (Financial Risk Manager)}: a globally recognized certification in financial risk management, focusing on financial risk assessment tools and their applications.
    
    \item ** \textbf{CLIQ (Chinese Life
Insurance Qualification)}: a professional qualification exam for Chinese insurance practitioners, covering insurance basics, risk management, insurance products, and related laws and regulations.
    
    \item *** \textbf{FundPQ (Fund Practitioner Qualification)}: a professional certification exam for Chinese fund practitioners, covering fund laws and regulations, professional ethics, business standards, and basic knowledge of securities investment funds or private equity funds.
    
    \item *** \textbf{FuturesPQ (Futures Practitioner Qualification)}: an entry-level certification exam for Chinese futures practitioners, covering ``Futures Basics'', ``Futures Laws and Regulations'', and ``Futures Investment Analysis''.
    
    \item ** \textbf{Economist}: a professional qualification exam for economics, focusing on professional knowledge and practical skills in economics.
    
    \item *** \textbf{SPQ (Securities Practitioner Qualification)}: a qualification exam for the Chinese securities industry, covering basic knowledge of financial markets and securities market laws and regulations.
    
    \item *** \textbf{CCBP (Certification of China Banking Professional)}: a qualification certification for Chinese banking practitioners, including two subjects: banking laws and regulations with comprehensive capability, and professional banking practice.
\end{itemize}

\begin{figure}
 \centering
\includegraphics[width=8cm]{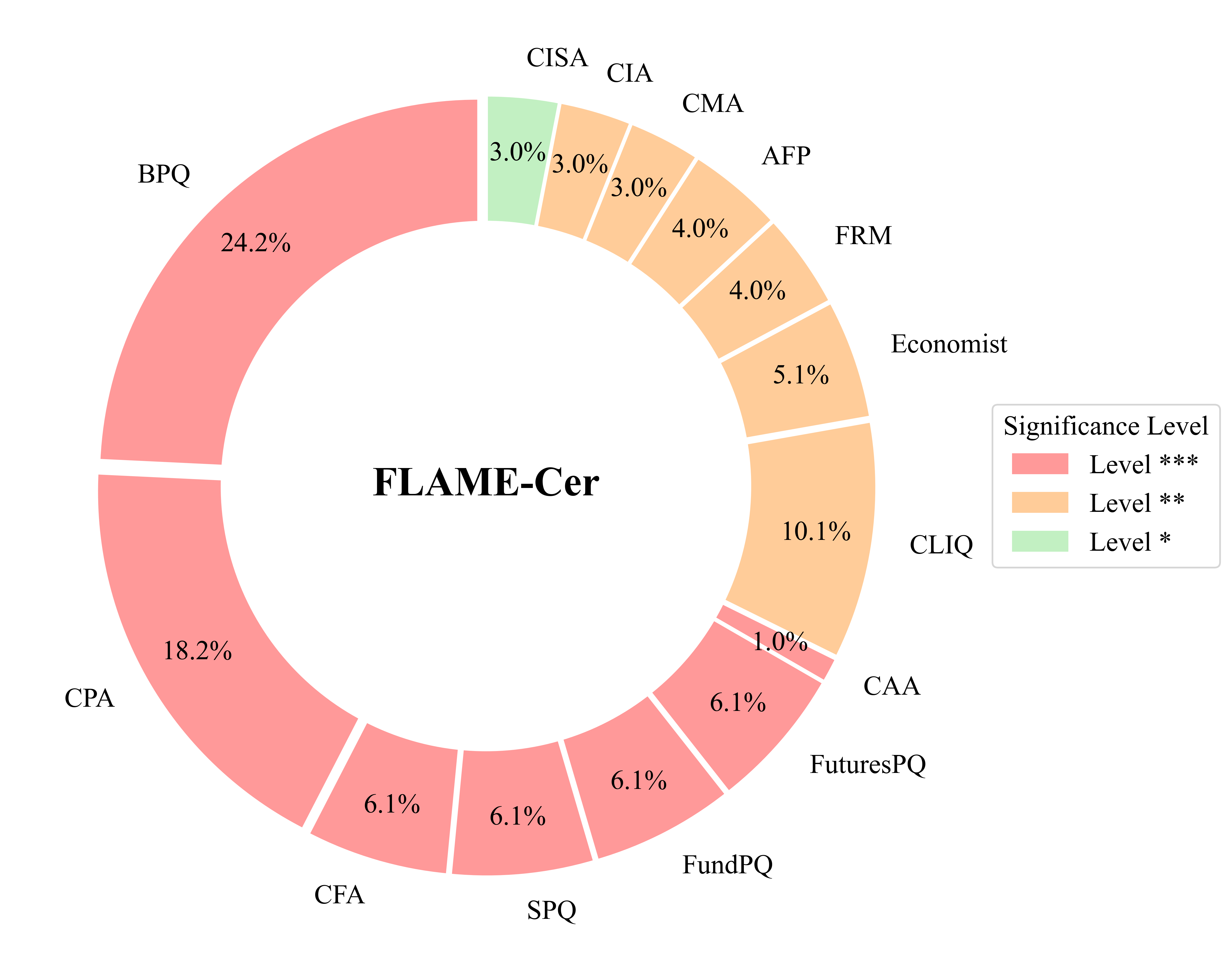}
\caption{The composition of FLAME-Cer benchmark.}
\label{fig:FLAME-Cer}
\end{figure}

We also visualize the data proportion of Flame-Cer in Figure~\ref{fig:FLAME-Cer}.

\subsection{Evaluation Metrics of FLAME-Cer}

For FLAME-Cer, accuracy is used as the evaluation metric. This is measured by calculating the percentage of correct answers the model achieves on the test set.

FLAME-Cer effectively assesses the mastery of financial professional knowledge by LLMs, providing a reliable reference for financial institutions in selecting and applying these models.

\subsection{FLAME-Sce: Financial Scenario Application Benchmark}

FLAME-Sce is a professional financial scenario evaluation benchmark designed to comprehensively assess the practical application capabilities of LLMs in the financial domain through a systematic hierarchical framework and rigorous evaluation standards.

FLAME-Sce encompasses 10 primary core business scenarios, including Financial Knowledge and Theory, Financial Compliance, Financial Document Generation, Financial Intelligent Customer Services, Financial Risk Control, Financial Document Processing, Financial Analysis and Research, Financial Investment and Wealth Management, Financial Marketing, and Financial Data Processing. In these primary scenarios, there are 21 secondary scenarios, and nearly 100 tertiary application tasks, offering a wide variety of task types, including term explanations, calculation problems, short-answer questions, material analysis, and comprehensive argumentation tasks. The final FLAME-Sce contains over 5,000 high-quality evaluation questions, covering the full spectrum from fundamental financial knowledge to complex business analysis.

We present a detailed introduction to each secondary scenario below.

\begin{itemize}
    \item \textbf{Financial Knowledge and Theory}
    \begin{itemize}
        \item \textbf{Basic Financial Knowledge}: understanding specialized terminology, fundamental theories, and historical events in the financial domain.
        \item \textbf{Financial Products / Business Knowledge}: understanding and applying knowledge related to product descriptions and institutional business processes in banking, insurance, and securities industries.
    \end{itemize}

    \item \textbf{Financial Compliance}
    \begin{itemize}
        \item \textbf{Financial Policies and Regulations}: understanding and applying financial policies, laws, and regulations issued by various regulatory authorities.
    \end{itemize}

    \item \textbf{Financial Document Generation}
    \begin{itemize}
        \item \textbf{Financial Report Generation}: writing common financial materials, including but not limited to financial contracts, due diligence reports, and research reports, based on various needs.
        \item \textbf{Financial Text Summarization}: extracting key information and core content from financial texts and presenting them in a concise and accurate manner.
    \end{itemize}

    \item \textbf{Financial Intelligent Customer Service}
    \begin{itemize}
        \item \textbf{Customer Intent Recognition}: analyzing communication records to understand and summarize customer intentions in financial scenarios.
        \item \textbf{Customer Behavior Analysis}: analyzing customer behavior inclinations from communication records to determine repayment willingness or potential risks.
        \item \textbf{Business Opportunity Identification}: identifying and evaluating potential business opportunities from customer communication records.
        \item \textbf{Customer Service Quality Inspection}: evaluating the service quality of customer service representatives based on communication records.
        \item \textbf{Customer Service Script Generation}: generating customer service responses tailored to specific consultation scenarios in banking, insurance, and securities industries.
    \end{itemize}

    \item \textbf{Financial Risk Control}
    \begin{itemize}
        \item \textbf{Identity Verification}: verifying the authenticity and legitimacy of customer identities in various financial transaction scenarios.
        \item \textbf{Transaction Safety Assessment}: verifying whether customers are aware of specified transactions based on communication records.
        \item \textbf{Fraud Detection}: identifying potential risks or fraudulent activities in financial transactions based on communication records.
        \item \textbf{Public Opinion Monitoring and Analysis}: collecting and analyzing public opinions, attitudes, and
        emotions on financial events to assess their potential impact on financial products,
        services, or market trends.
    \end{itemize}

    \item \textbf{Financial Document Processing}
    \begin{itemize}
        \item \textbf{Financial Document Information Extraction}: automatically identifying and extracting key information from financial documents and outputting structured data.
        \item \textbf{Financial Conversation Analysis}: exploring customer needs and preferences through direct or indirect dialogue records.
    \end{itemize}

    \item \textbf{Financial Analysis and Research}
    \begin{itemize}
        \item \textbf{Industry Research}: analyzing and interpreting various industry research reports, including identifying key data, trend analysis, market forecasts, and policy impacts.
    \end{itemize}

    \item \textbf{Financial Investment and Wealth Management}
    \begin{itemize}
        \item \textbf{Investment Preparation}: analyzing customers' risk tolerance and investment preferences based on their basic information and transaction behaviors to build customer profiles.
        \item \textbf{Investment Advisory Assistance}: providing personalized asset allocation recommendations based on modern portfolio theory and the investor's risk preferences and return objectives.
    \end{itemize}

    \item \textbf{Financial Marketing}
    \begin{itemize}
        \item \textbf{Marketing Content Generation}: generating personalized marketing scripts for promoting different financial products to various customers.
    \end{itemize}

    \item \textbf{Financial Data Processing}
    \begin{itemize}
        \item \textbf{Financial Data Calculation}: analyzing and processing financial data to calculate various key financial ratios and metrics.
    \end{itemize}
\end{itemize}

The evaluation scenarios in our FLAME-Sce are derived from the real business needs of financial institutions and the evaluation content is regularly updated to reflect market changes and emerging business needs.

\subsection{Evaluation Metrics of FLAME-Sce}

FLAME-Sce employs a multi-dimensional manual evaluation approach, with differentiated evaluation dimensions and weight configurations tailored to the characteristics of each scenario.

\subsubsection{Basic Evaluation Dimensions}

The FLAME-Sce evaluation framework is designed with a comprehensive set of basic evaluation dimensions to assess the performance of large models in practical financial scenarios. These dimensions are distilled from the common requirements across various financial application scenarios and are flexibly adjusted and weighted based on the characteristics of specific scenarios during evaluation.

\begin{itemize}
    \item \textbf{Accuracy}
    \begin{itemize}
        \item \textit{Information Accuracy}: Assesses the correctness of financial concepts, professional terms, data, and factual statements, as well as their alignment with real-world scenarios.
        \item \textit{Analytical Accuracy}: Evaluates the rigor of reasoning processes and the sufficiency of evidence supporting derived conclusions.
    \end{itemize}

    \item \textbf{Completeness}
    \begin{itemize}
        \item \textit{Content Coverage}: Examines whether the response fully addresses all key aspects of the question, including essential background information, core elements, and examples.
        \item \textit{Depth and Breadth}: Assesses the depth of analysis and the extensiveness of related knowledge points.
    \end{itemize}

    \item \textbf{Compliance}
    \begin{itemize}
        \item \textit{Regulatory Compliance}: Evaluates adherence to financial laws, regulations, and industry standards.
        \item \textit{Risk Control}: Examines the inclusion of adequate risk warnings, identification of potential risk points, and provision of risk mitigation suggestions.
    \end{itemize}

    \item \textbf{Relevance}
    \begin{itemize}
        \item \textit{Answer Relevance}: Assesses whether the content of the response directly addresses the core of the question without deviating from the topic.
        \item \textit{Intent Alignment}: Evaluates whether the provided information and suggestions align with the user's actual intent.
    \end{itemize}

    \item \textbf{Professionalism}
    \begin{itemize}
        \item \textit{Financial Expertise}: Assesses the application of financial theories, understanding of industry rules, and ability to perform professional analysis.
        \item \textit{Industry Insights}: Evaluates the grasp of market conditions, industry trends, and the ability to provide comprehensive analysis and professional recommendations.
    \end{itemize}

    \item \textbf{Clarity}
    \begin{itemize}
        \item \textit{Clear Expression}: Assesses the clarity of language, structural organization, and logical coherence.
        \item \textit{Ease of Understanding}: Evaluates the simplification of professional content and the inclusion of necessary explanations to facilitate understanding.
    \end{itemize}

    \item \textbf{Practicality}
    \begin{itemize}
        \item \textit{Practical Guidance}: Assesses the feasibility and practical value of the solutions provided by the model.
        \item \textit{Business Relevance}: Evaluates the alignment of the model's output solutions with real-world financial business practices.
    \end{itemize}

    \item \textbf{Instruction Compliance}
    \begin{itemize}
        \item Assesses the model's completeness in adhering to the instructions, including meeting specific requirements on output format, style, and content scope.
    \end{itemize}
\end{itemize}

\subsubsection{Scenario-Specific Evaluation Dimensions and Weight Configurations}

FLAME-Sce establishes customized evaluation dimension combinations and weight configurations for nearly 100 detailed tertiary financial application tasks, ensuring that the evaluation standards align closely with the characteristics of each scenario. Below are specific examples:

\begin{itemize}
    \item \textbf{Financial Material Generation - Financial Text Summarization - Meeting Minutes Summarization}
    \begin{itemize}
            \item \textbf{Accuracy (30\%)}: Assesses whether the summary covers all key information and decisions from the meeting, aligns fully with the meeting content, and avoids misunderstandings or misrepresentations.
            \item \textbf{Completeness (30\%)}: Evaluates whether the summary reflects the main topics, discussion points, and outcomes of the meeting comprehensively and retains critical details for understanding the context and rationale.
            \item \textbf{Clarity (10\%)}: Checks if the summary uses clear and simple language, avoids unnecessary jargon, and organizes content into distinct sections for easy navigation.
            \item \textbf{Practicality (10\%)}: Determines if the summary includes clear action plans, responsible parties, and deadlines to support follow-ups, and provides sufficient information for decision-making by management.
            \item \textbf{Instruction Compliance (20\%)}: Assesses whether the summary adheres strictly to instruction requirements, including any constraints on output format or structure.
        \end{itemize}

    \item \textbf{Financial Investment and Wealth Management - Investment Preparation - Customer Profile Analysis}
    \begin{itemize}
            \item \textbf{Accuracy (30\%)}: Checks whether all extracted customer information aligns perfectly with the original data, without any errors, and whether analytical conclusions are fact-based, rigorous, and precise.
            \item \textbf{Completeness (30\%)}: Evaluates whether the analysis covers all key dimensions, including basic characteristics, behavioral traits, and needs, and whether it provides valuable insights.
            \item \textbf{Logicality (20\%)}: Assesses whether the analytical process is clearly structured, well-organized, and logically consistent, with seamless transitions and well-supported conclusions.
            \item \textbf{Instruction Compliance (20\%)}: Examines whether the analysis fully complies with the instructions, adhering to any constraints such as output format requirements.
    \end{itemize}

    \item \textbf{Financial Compliance - Financial Policies and Regulations - Tax Policy Q\&A}
    \begin{itemize}
            \item \textbf{Accuracy (40\%)}: Evaluates whether the information in the response aligns with the content of tax policy documents, avoids misleading statements, and ensures the accuracy of any statistical data or specific figures provided.
            \item \textbf{Relevance (20\%)}: Checks whether the answer focuses entirely on the specific tax policy queried by the user, addressing all key points and avoiding irrelevant content, while aligning with the user's application context.
            \item \textbf{Completeness (20\%)}: Assesses whether the response fully covers the regulations involved in the question, includes all critical information, and provides sufficient detail and background for explanations where necessary.
            \item \textbf{Instruction Compliance (20\%)}: Ensures the response strictly follows instruction requirements, including any constraints on output format or structure.
        \end{itemize}
\end{itemize}

\subsubsection{Scoring Strategy}

FLAME-Sce employs a two-step scoring mechanism that combines dimension-specific scoring with weighted calculations to provide a quantitative evaluation of model performance across various scenarios:

\begin{itemize}
    \item \textbf{Step 1: Dimension Scoring}: Scores range from -1 to 5. A score of 5 represents optimal performance in the given dimension. A score of -1 indicates a severe violation of the basic requirements for the dimension (e.g., compliance violations).

    \item \textbf{Step 2: Weighted Calculation}: Scores are weighted based on the scenario-specific dimension weight configurations.
        \[
        \text{Final Score} = \sum (\text{Dimension Score} \times \text{Dimension Weight})
        \]
\end{itemize}

For simple scenarios, e.g., financial term explanations, a score of ≥3 is deemed usable. For complex scenarios, e.g., risk control, a score of ≥4 is deemed usable. If any dimension receives a score of -1, the overall result is deemed unusable.

This scoring strategy ensures that the model's performance receives accurate and fair evaluations in each specific scenario.

\section{Experiments}

\subsection{Baselines \& Settings}
We conducted a comprehensive evaluation of representative LLMs under zero-shot setting, including Qwen2.5-72B-Instruct~\footnote{\url{https://huggingface.co/Qwen/Qwen2.5-72B-Instruct}}, a widely recognized model in the open-source ecosystem; GPT-4o~\footnote{\url{https://openai.com/index/hello-gpt-4o/}}, an industry-leading LLM; XuanYuan3-70B-Chat~\footnote{\url{https://huggingface.co/Duxiaoman-DI/XuanYuan-70B}}, an open-source LLM tailored for the financial domain in the Chinese context; ERNIE-4.0-Turbo-128K~\footnote{\url{https://cloud.baidu.com/}} and GLM-4-PLUS~\footnote{\url{https://open.bigmodel.cn/}}, two renowned closed-source LLMs in China; and Baichuan4-Finance~\footnote{\url{https://commercial-platform-cdn.oss-cn-beijing.aliyuncs.com/openapi/20241218/Baichuan4_Finance_Technical_Report.pdf}}, a model specifically enhanced for financial applications.

\subsection{Evaluation Results on FLAME-Cer}

The performance comparison of LLMs on the FLAME-Cer benchmark is shown in Table~\ref{tab:performance_flame_cer}. We could find that Baichuan4-Finance leads with an average accuracy rate of 93.62\%, followed by Qwen2.5-72B-Instruct at 88.24\%. GPT-4o, GLM-4-PLUS, and ERNIE-4.0-Turbo-128K demonstrate similar performance levels, with accuracy rates ranging between 77\% and 81\%.

Examining the performance across specific exam subjects, the models generally perform well in the Fund, Securities, and Banking Qualification Exams, with most LLMs achieving accuracy rates above 80\%. The CAA (China Actuary Association) Exam poses a significant challenge for all LLMs, with accuracy rates remaining low; even the best-performing model, Baichuan4-Finance, achieves only 66.51\%. Notably, Baichuan4-Finance surpasses 95\% accuracy in several exam subjects, excelling particularly in the Fund and Securities Qualification Exams with accuracy rates of 97.93\% and 97.60\%, respectively.

\begin{table}[t]
\centering
\caption{The performance of LLMs on FLAME-Cer benchmark.}
\resizebox{\linewidth}{!}{
\begin{tabular}{ccccccc}
    \toprule
    \textbf{FLAME-Cer} & \textbf{GPT-4o} & \textbf{\makecell{ERNIE-4.0-\\Turbo-128K}} & \textbf{\makecell{GLM-4-\\PLUS}} & \textbf{\makecell{Qwen2.5-72B-\\Instruct}} & \textbf{\makecell{XuanYuan3-70B-\\Chat}} & \textbf{Baichuan4-Finance} \\
    \midrule
    \textbf{AFP} & 74.19 & 72.93 & 79.95 & 79.37 & 64.09 & \textbf{90.08} \\
  \textbf{CFA} & \textbf{86.97} & 74.44 & 81.45 & 82.92 & 63.88 & 85.59 \\
   \textbf{CAA} & 44.65 & 48.37 & 43.26 & 55.81 & 31.63 & \textbf{66.51} \\
   \textbf{CIA} & 84.46 & 77.94 & 83.46 & 86.69 & 70.53 & \textbf{92.59} \\
  \textbf{CISA} & 86.72 & 77.94 & 83.96 & 86.83 & 70.31 & \textbf{95.76} \\
  \textbf{CMA} & 83.21 & 73.68 & 76.69 & \textbf{86.10} & 65.71 & \textbf{86.10} \\
   \textbf{CPA} & 68.67 & 70.18 & 78.95 & 85.31 & 68.64 & \textbf{93.16} \\
  \textbf{FRM} & 74.94 & 67.92 & 76.69 & 76.87 & 55.46 & \textbf{82.87} \\
  \textbf{CLIQ} & 81.20  & 79.95 & 81.70  & 86.61 & 69.10  & \textbf{93.82} \\
  \textbf{FundPQ} & 82.96 & 84.21 & 84.71 & 94.61 & 76.73 & \textbf{97.93} \\
  \textbf{FuturesPQ} & 75.94 & 78.45 & 82.46 & 90.23 & 73.74 & \textbf{96.54} \\
  \textbf{Ecomonist} & 80.45 & 87.97 & 90.48 & 93.41 & 82.35 & \textbf{95.78} \\
   \textbf{SPQ} & 76.69 & 84.21 & 88.72 & 94.33 & 78.74 & \textbf{97.60} \\
  \textbf{CCBP} & 78.70  & 86.97 & 86.47 & 92.72 & 79.26 & \textbf{96.42} \\
\midrule   \textbf{Average} & 78.23 & 77.03 & 81.17 & 88.24 & 72.07 & \textbf{93.62} \\
    \bottomrule
    \end{tabular}}%
\label{tab:performance_flame_cer}
\end{table}

\begin{table}[t]
\centering
\caption{The performance of LLMs on FLAME-Sce benchmark.}
\resizebox{\linewidth}{!}{
    \begin{tabular}{ccccccc}
    \toprule
    \textbf{FLAME-Sce} & \textbf{GPT-4o} & \textbf{\makecell{ERNIE-4.0-\\
    Turbo-128K}} & \textbf{\makecell{GLM-4-\\PLUS}} & \textbf{\makecell{Qwen2.5-72B-\\Instruct}} & \textbf{\makecell{XuanYuan3-70B-\\Chat}} & \textbf{Baichuan4-Finance} \\
    \midrule
    \textbf{Knowledge and Theory } & 89.14 & 87.67 & 87.06 & 87.75 & 76.94 & \textbf{91.17} \\
 \midrule
    \textbf{Compliance} & 80.27 & 80.27 & 86.95 & 83.61 & 40.13 & \textbf{87.24} \\
 \midrule
    \textbf{Document Generation} & 60.01 & 58.89 & 61.11 & 58.89 & 41.11 & \textbf{70.03} \\
 \midrule
    \textbf{\makecell{Intelligent Consumer\\ Services}} & 83.52 & 79.59 & 79.78 & 83.18 & 62.37 & \textbf{86.92} \\
 \midrule
    \textbf{Risk Control} & 82.94 & 75.86 & 80.3  & 81.99 & 63.93 & \textbf{85.36} \\
 \midrule
    \textbf{Document Processing} & 83.51 & 80.22 & 65.88 & 80.18 & 65.91 & \textbf{86.77} \\
 \midrule
    \textbf{Data Processing} & 86.43 & 89.99 & 79.87 & 91.61 & 71.92 & \textbf{91.71} \\
 \midrule
    \textbf{Analysis and Research} & 45.45 & \textbf{48.48} & 42.42 & 42.42 & 33.33 & 45.45 \\
 \midrule
    \textbf{Marketing} & 57.58 & \textbf{93.94} & 75.76 & 54.55 & 30.3  & 78.79 \\
 \midrule
    \textbf{\makecell{Investment \& \\ Wealth Management}} & 85.71 & 71.43 & \textbf{89.29} & 85.71 & 67.86 & \textbf{89.29} \\
    \midrule
    \textbf{Average} & 79.88 & 78.01 & 76.92 & 79.18 & 61.34 & \textbf{84.15} \\
    \bottomrule
    \end{tabular}%
}
\label{tab:performance_flame_sce}
\end{table}

\subsection{Evaluation Results on FLAME-Sce}

The performance comparison of LLMs on FLAME-Sce benchmark is shown in Table~\ref{tab:performance_flame_sce}. Overall, Baichuan4-Finance leads with a usability rate of 84.15\%. GPT-4o and Qwen2.5-72B-Instruct exhibit similar performance levels, around 79\%. ERNIE-4.0-Turbo-128K and GLM-4-PLUS also perform well, achieving rates between 76\% and 78\%. XuanYuan3, with a usability rate of 61.34\%, shows room for improvement in financial scenario applications.

In terms of specific scenarios, most LLMs perform well in financial knowledge theory and financial data computation, with usability rates generally exceeding 80\%. However, they tend to score lower in financial analysis and financial document generation, highlighting areas for improvement across all LLMs. Baichuan4-Finance consistently demonstrates high performance across most scenarios, excelling in financial intelligent customer service and financial risk control (85–87\%).Qwen2.5-72B-Instruct achieves an impressive 91.61\% in data computation scenarios, while ERNIE-4.0-Turbo-128K stands out in financial marketing scenarios with a rate of 93.94\%. GLM-4-PLUS showcases its strengths in financial compliance (86.95\%) and financial investment and wealth management (89.29\%).

\subsection{Case Study}

\subsubsection{Case Study of FLAME-Cer}
We provide one case example for each certification in FLAME-Cer as follows.

\begin{itemize}

    \item \textbf{AFP (Associate Financial Planner, 金融理财师)}\\ 现年25岁的王晓峰，计划工作10年后购买总价80万元的两居室，再10年后换购总价120万元的三居室。假设换房时，旧房按原价出售，投资报酬率为10\%，房贷利率为6\%，贷款20年，贷款70\%，则王晓峰从25～35岁间配置在购房上的年储蓄额应达\underline{\hspace{1cm}}元，36～45岁应提高至\underline{\hspace{1cm}}元，46～65岁间应提高至\underline{\hspace{1cm}}元，才能完成一生的房涯规划。（A）\\
A.15059；15223；44642 \hspace{1cm}B.15223；15059；44642 \\C.15223；17325；45620 \hspace{1cm}D.15059；17325；45620
    
    \item \textbf{CAA (China Actuarial Association, 中国精算师)}\\ 一种5年期债券的现货价格为950元，该债券1年期远期合约的交割价格为960元，该债券在6个月末和12个月末都将收到50元的利息，且第二次付息日在远期合约交割日之前。假设6个月期和22个月期的无风险年利率（连续复利）分别为9\%和10\%，则该远期合约空头的价值为（B）元。\\
A.12.43 \hspace{1cm} B.11.68 \hspace{1cm} C.14.43 \hspace{1cm} D.-14.43 \hspace{1cm} E.-11.68
    
    \item \textbf{CFA (Chartered Financial Analyst, 特许金融分析师)}\\ Marc Davidson, CFA, works as a trust specialist for Integrity Financial. Davidson starts a part-time consulting business providing advice to trustees for a fee. He conducts this business on his own time and therefore did not notify Integrity Financial of his consulting. Davidson asks his assistant to compile a list of Integrity's clients and their contact information. The following month, Davidson is offered a similar role at Integrity's largest competitor, Legacy Trust Services, Inc. After he begins working at Legacy, his new manager arranges for him to meet with a number of prospective clients, many of whom are clients of Integrity. After meeting with Davidson, a number of former Integrity clients decide to transfer their business to Legacy. Did Davidson's action violate the Code and Standards? (B)\\
A.Yes, both Davidson's part-time consulting business and his meetings with Integrity clients are violations of the Standards. \\B.Yes, Davidson's part-time consulting business is a violation of the Standards. \\C.No.
    
    \item \textbf{CIA (Certified Internal Auditor, 国际注册内部审计师)}\\ 下列选项中，关于COSO企业全面风险管理框架和COSO内部控制框架的差异的说法不正确的是：(C)\\
A.全面风险管理框架在涵盖内部控制框架的基础上还增加了内容。 \\B.全面风险管理区分了风险和机会而内部控制框架仅仅关注风险的不利方面。\\ C.全面风险管理框架中的内部环境要素与内部控制框架中的控制环境是同样的含义。\\ D.引入风险偏好、风险容忍度等概念和方法后，全面风险管理框架有利于企业的发展战略和风险偏好相一致。
    
    \item  \textbf{CISA (Certified Information Systems Auditor, 国际注册信息系统审计师)}\\ 对信息系统审计师而言，验证关键生产服务器是否在运行供应商发布的最新安全更新的最佳方法是以下哪一项？(D)\\
    A.确保在关键生产服务器上已启用自动更新。\\
    B.在生产服务器样机上人工验证已应用的修补程序。\\
    C.审查关键生产服务器的变更管理日志。\\
    D.在生产服务器上运行自动化工具，以验证修补程序的安全性。
    
    \item \textbf{CMA (Certified Management Accountant, 注册管理会计)}\\ 在物价上涨期间，Mariposa公司为何选择先进先出法？(B)\\
A.记录每件贵重商品的准确利润。 \hspace{1cm} B.记录最高毛利润。\\C.使用更高的销售成本。 \hspace{1cm} D.使货物对买方更有吸引力。
    
    \item \textbf{CPA (Certified Public Accountant, 注册会计师)}\\ 如果集团管理层拒绝向组成部分管理层通报可能对组成部分财务报表产生重要影响的事项，集团项目组应当 (B)。  \\
A.视为审计范围受限，考虑对审计意见的影响。 \hspace{1cm}B.与集团治理层进行讨论。 \\
C.直接与组成部分管理层沟通。  \hspace{1cm} D.与组成注册会计师进行讨论。\\
    
    \item  \textbf{FRM (Financial Risk Manager, 金融风险管理师)}\\ The VaR of a portfolio at 95\% confidence level is 15.2. If the confidence level is raised to 99\% (assuming a one-tailed normal distribution), the new value of VaR will be closest to: (D)\\
A.10.8 \hspace{1cm} B.5.2 \hspace{1cm} C.18.1 \hspace{1cm} D.21.5\\
    
    \item  \textbf{CLIQ (Chinese Life
Insurance Qualification, 中国人身保险从业人员资格)}\\ 现金流量表是指反映公司在一定会计期间现金和现金等价物流入和流出的报表。关于现金流量表的表述中，错误的是: (A)\\
A.从编制原则上看，现金流量表按照权责发生制原则编制。\\
B.在现金流量表中，现金及现金等价物被视为一个整体，公司现金形式的转换不会产生现金的流入和流出。\\
C.现金流量表中的经营活动是指公司投资活动和筹资活动以外的所有交易和事项。\\
D.现金流量表中的投资活动，既包括实物资产投资，也包括金融资产投资。\\
    
    \item \textbf{FundPQ (Fund Practitioner Qualification, 基金从业资格证)}\\ 主动比重是一个相对于业绩比较基准的风险指标，用来衡量投资组合相对于基准的\underline{\hspace{1cm}}。(A)\\
A.偏离程度 \hspace{1cm} B.风险高度 \hspace{1cm} C.比重率 \hspace{1cm} D.期望收益
    
    \item \textbf{FuturesPQ (Futures Practitioner Qualification, 期货从业资格证)}\\ 国债期货价格和国债市场利率的关系是\underline{\hspace{1cm}}。 (A)\\
A.反向变动关系 \hspace{1cm} B.同向变动关系 \hspace{1cm}C.无相关关系 \hspace{1cm} D.线性关系\\
    
    \item \textbf{Economist, 经济师}\\ 通常情况下，直接影响利率水平的主要因素有\underline{\hspace{1cm}}。(BCDE)\\
A.政治制度 \hspace{1cm} B.平均利润率 \hspace{1cm} C.货币资金供求状况 \\
D.通货膨胀率 \hspace{1cm} E.中央银行货币政策\\
    
    \item \textbf{SPQ (Securities Practitioner Qualification， 证券从业资格证)}\\ 封闭式基金的基金份额在基金存续期间，基金份额持有人\underline{\hspace{1cm}} 。(A)\\
A．不得申请赎回 \hspace{1cm} B．在任何场所申请赎回 \\
C．可以在基金合同约定的时间和场所申请赎回 \hspace{1cm}D．可在任何时间申请赎回\\
    
    \item \textbf{CCBP (Certification of China Banking Professional)}\\直接税是税收负担不能由纳税人转嫁出去，必须由自己负担的税种，下列不属于直接税的是\underline{\hspace{1cm}}。(C)\\
A.企业所得税 \hspace{1cm} B.个人所得税\hspace{1cm} C.消费税 \hspace{1cm} D.房产税
\end{itemize}

\subsubsection{Case Study of FLAME-Sce}

We present a case example for each primary financial scenario in FLAME-Sce as follows.

\begin{itemize}
    \item \textbf{Financial Knowledge and Theory, 金融知识及理论}  \\
   以下是一个金融行业常用的术语（黑话），请对这个术语进行解释：梭哈\\

    \item \textbf{Financial Compliance, 金融合规}  \\
    银行在发现哪些情况时，应将单位银行结算账户的网上银行转账功能关闭，并要求存款人到银行网点柜台办理转账业务？ \\
    
    \item \textbf{Financial Document Generation, 金融材料生成}  \\
    你是一位资深法律顾问，具备丰富的金融知识及法律知识，擅长起草银行业务中涉及的各类合同。请你根据【合同需求】，参考【合同模版】中的内容起草一份外汇借款合同，以便保障借贷双方的权益，确保合同条款详尽、合法合规。\\

\# 输出要求：\\
1. 严格按照【合同模板】进行填写，确保所有条款的准确性，确保内容合法且结构完整；
2. 对引用的法律法规进行检查，去掉已经失效的法律法规，合同中的法律条款、利息需符合最新的中国法律；
3. 落款日期按照生成文本的日期；
4.空白的地方要根据理解填充
5. 使用XXX和XX银行代替实际人名和公司名称；
6. 输出格式为Markdown。

\# 合同需求：\\
流动资金贷款
合同双方信息
借款单位：XX基础设施建设有限公司（以下简称甲方）
贷款银行：中国工商银行（以下简称乙方）
借款金额：1亿美元（USD）。
借款期限：10年，从2024年8月1日至2034年7月31日。
贷款年利率：固定利率3\%。
利息计收：每年计收一次，结息日为每年的借款对应日。
用款计划：甲方分五年等额提取借款，每年提取2000万美元，分别在合同签订后的第一至第五年的对应日提取。
还本付息计划：甲方在借款期限的最后五年内每年偿还本金和利息，每年偿还本金的五分之一及相应利息。

\# 合同模板\\
以下是合同模版
[context]\\

    \item \textbf{Financial Intelligent Customer Service, 金融智能客服}  \\
    作为一名专业的银行客服，你的任务是根据客服-客户通话记录对客户来电意图进行准确分类。这项工作对于提升客户服务质量、优化产品策略和改进营销方案至关重要。\\

请按照以下分类标准对客户来电意图进行分类：\\
1. 存款产品：活期存款、定期存款、大额存单等
2. 贷款产品：个人贷款、房贷、车贷、信用贷款等
3. 保险业务：人寿保险、财产保险、健康保险等
4. 理财产品：基金、股票、债券、理财计划等
5. 养老产品：养老保险、养老金、养老理财等
6. 特色业务：外汇服务、信用卡业务、网上银行、手机银行等

要求：\\
1. 仔细阅读通话记录，识别关键信息
2. 如果一个通话涉及多个意图，请列出所有主要意图、次要意图，并按重要性排序
3. 给出你的分类结果，并提供简要的理由（不超过100字）
4. 如果无法确定意图，请在意图分类时标注为"未分类"并说明原因

输出格式：\\
- 主要意图：[分类结果]
- 次要意图（如有）：[分类结果]
- 分类理由：[简要解释]

请分析以下通话记录：
[context]

    \item \textbf{Financial Risk Control, 金融风控} \\ 
你是一名正在检查客户的账户安全的银行柜员，你需要根据提供的客服-客户通话记录，判断某笔交易是否是客户自己操作的。\\

判断规则：\\
1.若客户明确且肯定地表示交易是自己操作的，或能够清晰说明操作的原因、时间、地点等相关细节，判断为本人操作。
2.若客户明确且坚决地表示交易不是自己操作的，如“没有啊”、”不是我“，或能提供相应的不在场证明或合理的解释，判断为非本人操作。
3.若客户对于交易是否为本人操作的表述模糊不清，如“我不太确定”“可能是我，也可能不是”，则判断为”无法确定是否本人操作“。\\

请按照给定格式输出你的结论即可，不要输出其他内容。\\
给定格式：我的结论是：\\

客服-客户通话记录如下：[context]

    \item \textbf{Financial Document Processing, 金融材料处理}  \\
你是一位专业的财务数据分析师，负责从提供的已知的财务报表中抽取特定信息。你的任务是针对用户提出的问题，从财务报表中提取相关数据。\\
 
财务报表OCR文本：[context]\\
 
原则遵循：\\
1. 准确性：严格基于“财务报表”进行信息抽取，确保所有数据的准确无误。
2. 完整性：如果“财务报表”中包含用户问题所需的所有数据，则提供完整的答案；如果缺少信息，则在相应的字段中留空。
3. 输出格式：以JSON形式输出抽取的信息，确保易于阅读和理解。\\

用户问题：截止2024年3月底，淘宝和天猫集团的调整后EBITA是多少？\\

    \item \textbf{Financial Analysis and Research, 金融分析与研究}  \\
    角色\\
你是一位在金融行业深耕多年的资深分析师，目前你的主要工作是专注于对各类报告进行深入解读，致力于确保用户能够清晰、准确地把握和理解报告中的核心内容与信息。\\

指令要求：\\
1.细致研读以下企业研究报告，系统梳理全文的逻辑框架与核心要点。
2.全面搜索并整合该报告中涉及该企业发展趋势的所有相关信息，确保信息的完整性和准确性，避免任何遗漏。
3.剔除冗余及重复的观点，精炼并归纳出该企业明确的发展方向与趋势。
可分点论述该企业发展趋势：[发展趋势一]：[发展趋势二]：[发展趋势三]：\\

注意：你所输出的信息需经过精心编纂，确保内容的深度与广度在限定的1500字范围内得到最优展现，不容许有丝毫超出字数限制的冗余。\\

 报告内容：
[context]\\

    \item \textbf{Financial Marketing, 金融营销}  \\
   作为银行的金融顾问，你的目标是通过各种客户接触点，向小微企业主介绍我们的贷款产品。你的任务是让客户感受到产品的价值，并激发他们深入了解产品的兴趣。\\
 
产品概览：\\
名称：中银企E贷
特点：这是一款线上贷款产品，支持抵押和信用两种模式。信用贷额度最高可达100万元，无需抵押，适合信用良好的小微企业，抵押贷额度则最高可达1000万元。\\
 
营销框架：\\
1. 故事讲述：
   通过讲述一个相关企业如何通过使用我们的产品实现业务增长的故事来吸引客户。
2. 情景模拟：
   描述一个小微企业可能面临的具体资金挑战，以及该产品如何成为解决方案。
3. 价值强调：
   强调产品的主要价值主张，如快速审批流程、灵活的还款选项等。
4. 互动提问：
   通过提出问题来引导客户思考他们的资金需求，并自然过渡到产品介绍。\\
 
示例话术：\\
“我最近遇到了一个本地咖啡店老板，他们正计划扩展业务但遇到了资金瓶颈。通过我们的xx贷款产品，他们不仅迅速获得了所需的资金，而且还款计划也非常灵活。想象一下，如果您的业务面临类似的机遇，这款产品可能是您的理想选择。您是否曾经考虑过如何快速获得资金来支持您的业务增长？我们在这里可以帮助您探索更多可能性。”\\
 
注意事项：\\
1.情感连接：通过故事讲述和情景模拟，与客户建立情感上的联系。
2.解决方案导向：专注于展示产品如何解决实际问题，而不仅仅是产品特性。
3.互动性：通过提问和引导客户参与对话，增加互动性。
4.简洁性：保持话术简洁，避免过多技术术语，确保信息易于理解。\\

    \item \textbf{Financial Investment and Wealth Management, 金融投资与财富管理}  \\
    
    客户基本情况：[context]\\
作为一位资深的客户分析师,请你根据提供的客户基本信息(包括但不限于年龄、职业、收入水平、家庭状况、兴趣爱好等),全面分析并描述这位客户的生活状态和心理特征。\\

你的分析应涵盖以下几个方面:\\
1、客户对生活的整体满意度,包括主观幸福感和客观生活质量的评估。
2、客户当前最渴望得到或实现的事物,以及背后反映的需求和动机。
3、客户面临的主要烦恼和困惑,以及可能的原因和影响。
4、客户的生活态度和价值观,是积极还是消极,以及在日常生活中的具体表现。
5、客户的时间分配情况,包括工作、休闲、社交等方面的平衡。
6、客户的消费理念和行为模式,是偏向节俭还是享受型,在不同消费领域的表现。
7、客户的社会责任感和公益意识,以及参与社会活动的倾向。\\

在分析过程中,请注意以下要求:\\
- 保持客观中立,避免带有偏见或歧视性的表述。
- 运用心理学和社会学相关理论,对客户行为进行分析。
你的分析报告应当逻辑清晰,论述有据,语言简洁专业。如果客户基本情况里面没有相关信息，就不用分析。字数控制在500-600字之间。\\

    \item \textbf{Financial Data Calculation, 金融数据计算}  \\
    
    债券投资收益率：面值100元的债券，票面利率5\%，期限5年，每年付息一次，到期一次还本，若购买价格为90元，求该债券投资的到期收益率。
\end{itemize}

\section{Conclusion}

This paper introduces FLAME, a Chinese financial LLM evaluation system with two benchmarks: FLAME-Cer, covering 14 financial certifications with 16,000 manually reviewed questions, and FLAME-Sce, featuring nearly 100 financial application tasks across 10 primary and 21 secondary scenarios. Six representative LLMs, including GPT-4o and Baichuan4-Finance, were evaluated, with Baichuan4-Finance leading in most tasks. 

\bibliography{neurips_2024}
\bibliographystyle{plain}

\end{CJK}

\end{document}